# "Multisegmentation through wavelets: Comparing the efficacy of Daubechies vs Coiflets"

Madhur Srivastava, *Member, IEEE,* Yashwant Yashu, *Member, IETE,* Satish K. Singh, *Member, IEEE,* Prasanta K. Panigrahi

*Abstract---* In this paper, we carry out a comparative study of the efficacy of wavelets belonging to Daubechies and Coiflet family in achieving image segmentation through a fast statistical algorithm. The fact that wavelets belonging to Daubechies family optimally capture the polynomial trends and those of Coiflet family satisfy mini-max condition, makes this comparison interesting. In the context of the present algorithm, it is found that the performance of Coiflet wavelets is better, as compared to Daubechies wavelet.

Madhur Srivastava is final year B.TECH student in the Department of Electronics and Communication Engineering at Jaypee University of Engineering and Technology, Guna, India; e-mail: madhur.manas@gmail.com

Yashwant Yashu is final year B.TECH student in the Department of Electronics and Communication Engineering at Jaypee University of Engineering and Technology, Guna, India; e-mail: yashwantyashu.jiet@gmail.com

Satish K. Singh is Assistant Professor in the Department of Electronics and Communication Engineering at Jaypee University of Engineering and Technology, Guna, India; (Phone No. +91-9926467838) e-mail: satish432002@gmail.com

Prasanta K. Panigrahi is Professor in the Department of Physics at Indian Institute of Science Education and Research, Kolkata, India; e-mail: pprasanta@iiserkol.ac.in

I. INTRODUCTION

Thresholding of an image is done to reduce the storage space, increase the processing speed and simplify the manipulation as less number of levels are there compared to 256 levels of normal image. Primarily, thresholding are of two types – Bi-level and Multi-level [1].

Bi-level thresholding consists of two values – one below the threshold and another above it. While in Multilevel thresholding, different values are assigned between different ranges of threshold levels. Various thresholding techniques have been categorized on the basis of histogram shape, clustering, entropy and object attributes [2].

Wavelet Transform is very significant tool in the field of image processing. The wavelet transform of an image comprises four components – Approximation, Horizontal, Vertical and Diagonal. The process is recursively used in approximation component of wavelet transform for farther decomposition of image until only one coefficient is left in approximation part [3-5].

As is well known, Daubechies family is useful in extracting polynomial trends through their low-pass coefficients satisfying vanishing moments conditions:

$$\int_{-\infty}^{+\infty} x^n \psi_{j,k} dx = 0 \qquad (1)$$

This is due to the fact that, the wavelets of

$$\psi_{j,k} = 2^{j-1} \psi(2ix - k) \qquad (2)$$

For $n \leq N$, the values of *n* depend on the particular of this Daubechies basis makes them well suited for isolating smooth polynomial features in a given image. The Coiflet coefficient on the other hand, satisfies the mini-max condition, i.e., the maximum error in extracting local features is minimized in this basis set. Hence, it is worth comparing behavior of the corresponding wavelet at low-pass coefficients from the perspective of the proposed algorithm.

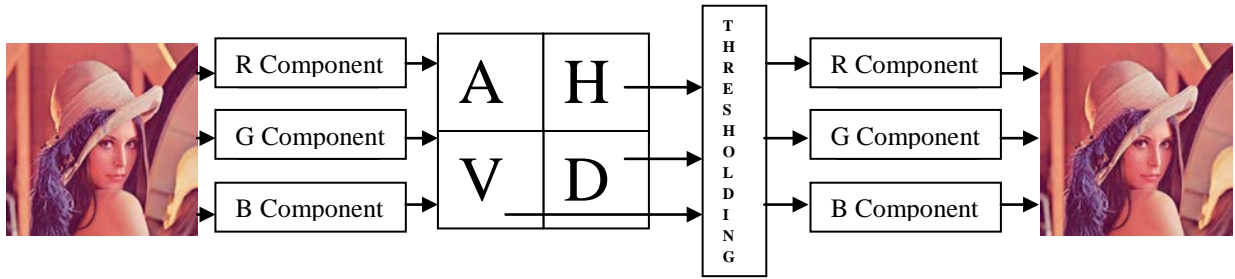

Fig. 1. Block diagram of the approach used.

## I. APPROACH

The thresholding applied in wavelet domain takes into account that majority of coefficients lie near to zero and coefficients representing large differences are few at the extreme ends of histogram plot for each horizontal, vertical and diagonal component. The coefficients with large differences represent most significant information of the image. Hence, the procedure provides for variable size segmentation with bigger block size around the mean, and having smaller blocks at the ends of histogram plot[6]. Following is the methodology used as shown in Fig. 1

1. Segregate the color image into its Red, Green and Blue components.
2. Take 2D-wavelet transform of each component at any level. Do the following for each Horizontal, Vertical and Diagonal part for every Red, Green and Blue component.
.
   - Threshold the coefficients using weighted mean and variance of each sub-band of histogram of coefficients.
   - Thresholding is done by having broader block size around mean while finer block size at the end of histogram.
3. Take inverse wavelet transform for each component.
4. Reconstruct the image by concatenating Red, Green and Blue components.

## III. RESULTS AND OBSERVATIONS

The proposed algorithm is tested on variety of standard images using Daubechies and Coiflet wavelets. The results of PSNR and size of reconstructed image at different threshold levels are shown in Table 1. The numbers of threshold levels taken are 3, 5 and 7. Figure 2 shows the graph of PSNR w.r.t threshold levels of the image of Lenna

Table 1. PSNR and size of reconstructed images using different Daubechies and Coiflet wavelets.

| Image Name | Threshold Level | | Wavelet Name | | | | | | | | |
|---|---|---|---|---|---|---|---|---|---|---|---|
| | | | dB2 | dB4 | dB6 | dB8 | coif1 | coif2 | coif3 | coif4 | coif5 |
| Lenna | 3 | PSNR(dB) | 34.45 | 35.19 | 35.52 | 35.71 | 34.50 | 35.23 | 35.48 | 35.61 | 35.69 |
| | | Size(kB) | 36.2 | 36.5 | 36.3 | 36.2 | 36.4 | 36.2 | 36.4 | 36.3 | 36.3 |
| | 5 | PSNR(dB) | 36.41 | 37.13 | 37.41 | 37.53 | 36.5 | 37.19 | 37.42 | 37.54 | 37.62 |
| | | Size(kB) | 36.2 | 36.5 | 36.3 | 36.3 | 36.3 | 36.3 | 36.4 | 36.4 | 36.4 |
| | 7 | PSNR(dB) | 36.79 | 37.5 | 37.74 | 37.88 | 36.84 | 37.53 | 37.76 | 37.89 | 37.97 |
| | | Size(kB) | 36.2 | 36.5 | 36.3 | 36.3 | 36.3 | 36.3 | 36.4 | 36.4 | 36.4 |
| Baboon | 3 | PSNR(dB) | 25.92 | 26.31 | 26.29 | 26.19 | 25.94 | 26.20 | 26.29 | 26.33 | 26.36 |
| | | Size(kB) | 74.4 | 74.2 | 74.2 | 74.3 | 74.4 | 74.4 | 74.3 | 74.3 | 74.2 |
| | 5 | PSNR(dB) | 27.06 | 27.56 | 27.44 | 27.40 | 27.13 | 27.41 | 27.50 | 27.55 | 27.58 |
| | | Size(kB) | 74.4 | 74.1 | 74.2 | 74.2 | 74.3 | 74.2 | 74.2 | 74.2 | 74.1 |
| | 7 | PSNR(dB) | 27.18 | 27.70 | 27.57 | 27.53 | 27.27 | 27.53 | 27.62 | 27.67 | 27.71 |

|  |  |  | | | | | | | | |
|---|---|---|---|---|---|---|---|---|---|---|
|  |  | Size(kB) | 74.3 | 74.1 | 74.1 | 74.1 | 74.2 | 74.2 | 74.1 | 74.2 | 74.1 |
| Pepper | 3 | PSNR(dB) | 30.63 | 33.87 | 31.61 | 31.25 | 31.48 | 31.63 | 31.70 | 31.75 | 31.77 |
|  |  | Size(kB) | 39.9 | 39.8 | 40.3 | 40.4 | 40.1 | 40.3 | 40.3 | 40.2 | 40.2 |
|  | 5 | PSNR(dB) | 34.12 | 35.83 | 34.61 | 34.30 | 33.98 | 34.41 | 34.55 | 34.60 | 34.62 |
|  |  | Size(kB) | 39.5 | 39.7 | 39.6 | 39.6 | 39.6 | 39.6 | 39.7 | 39.7 | 39.7 |
|  | 7 | PSNR(dB) | 34.56 | 36.26 | 34.92 | 34.58 | 34.25 | 34.73 | 34.88 | 34.93 | 34.95 |
|  |  | Size(kB) | 39.5 | 39.8 | 39.5 | 39.6 | 39.6 | 39.6 | 39.6 | 39.6 | 39.7 |

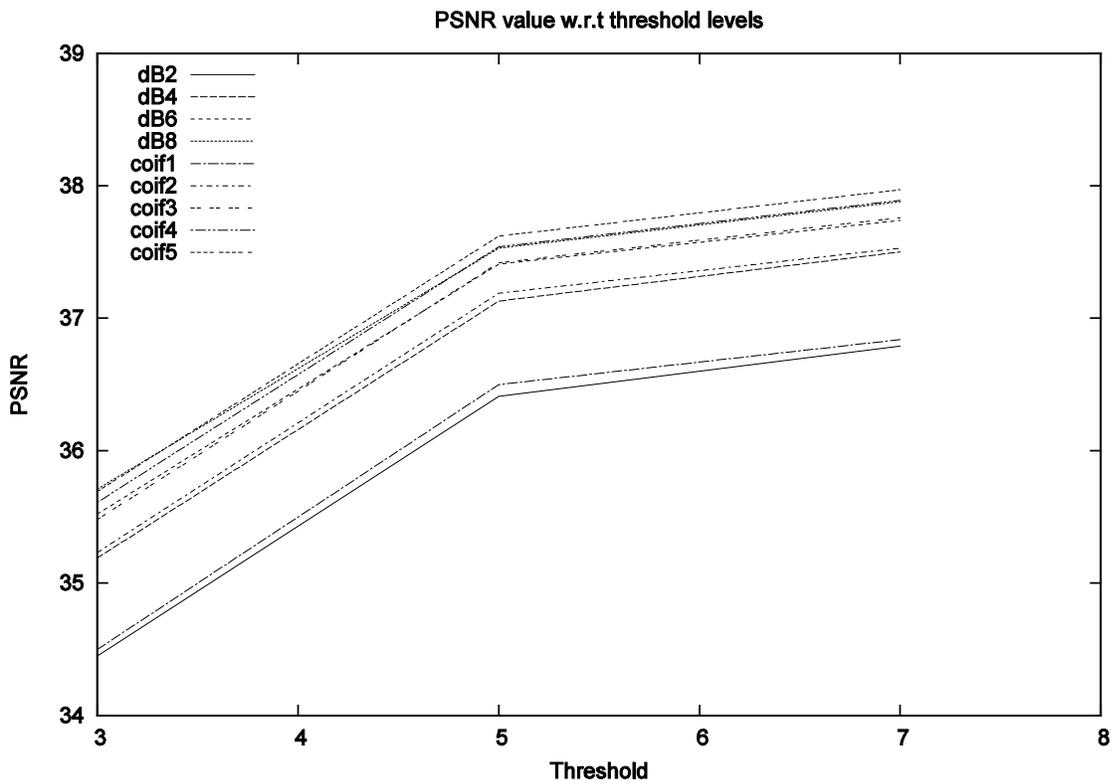

Fig. 2 Plot of PSNR vs Threshold levels thresholded using different wavelets of Lenna image

## IV. CONCLUSION

Thresholding performed by proposed algorithm gives better PSNR using coiflet wavelets compared to Daubechies wavelets while keeping the size of reconstructed image almost same. This is due to the unique property of coiflet wavelets satisfying the mini-max condition. Hence, it can be concluded that coiflet wavelets provides best and most desirable results during multilevel thresholding of image in wavelet domain. In future works, the proposed algorithm using coiflet wavelets can be used for image segmentation, object separation, image compression and image retrieval because only few coefficients of Horizontal, Vertical and Diagonal components represent the entire variation of image without deteriorating the quality.